\pdfoutput=1

\documentclass[11pt]{article}

\usepackage[final]{acl}

\usepackage{times}
\usepackage{latexsym}
\usepackage{multirow}

\usepackage[T1]{fontenc}

\usepackage[utf8]{inputenc}

\usepackage{microtype}

\usepackage{inconsolata}

\usepackage{graphicx}

\title{PatientDx: Merging Large Language Models for Protecting Data-Privacy in Healthcare}

\author{Jose G. Moreno$^{1}$ \hspace{1cm} Jesús Lovón-Melgarejo$^{1}$ \hspace{1cm} M'Rick Robin-Charlet$^{1,3}$  \\  \textbf{Christine Damase-Michel}$^{2}$ \hspace{1cm} \textbf{Lynda Tamine}$^{1}$\\ ${}^{1}$Université de Toulouse, IRIT UMR 5505, Toulouse, France\\
${}^{2}$Centre Hospitalier Universitaire de Toulouse \\ CERPOP INSERM UMR 1295 - SPHERE team, \\\ Faculté de Médecine Université de Toulouse, Toulouse, France\\
   ${}^{1}$\texttt{first.last@irit.fr \hspace{1cm} ${}^{2,3}$first.last@univ-tlse3.fr}}

\begin{document}
\maketitle
\begin{abstract}
Fine-tuning of Large Language Models (LLMs) has become the default practice for improving model performance on a given task. However, performance improvement comes at the cost of training on vast amounts of annotated data which could be sensitive leading to significant data privacy concerns. In particular, the healthcare domain is one of the most sensitive domains exposed to data privacy issues.   In this paper, we present \textit{PatientDx}, a framework of model merging that allows the design of effective LLMs for health-predictive tasks without requiring fine-tuning nor adaptation on patient data.  Our proposal is based on recently proposed techniques known as merging of LLMs and aims to optimize a building block merging strategy. \textit{PatientDx} uses a pivotal model adapted to numerical reasoning and tunes hyperparameters on examples based on a performance metric but without training of the LLM on these data.  Experiments using the mortality tasks of the MIMIC-IV dataset show improvements up to 7\% in terms of AUROC when compared to initial models. Additionally, we confirm that when compared to fine-tuned models, our proposal is less prone to data leak problems without hurting performance. Finally, we qualitatively show the capabilities of our proposal through a case study. Our best model is publicly available at \url{https://huggingface.co/Jgmorenof/mistral_merged_0_4}.

\end{abstract}

 \section{Introduction }
Recent breakthroughs made by the impressive capabilities of Large Language Models (LLMs) on one side, and the common practice of publishing them for a sharing purpose in the other side, have led to exploring their application to a wide range of applications and tasks.  Their strong performances heavily rely on their extremely large model architectures (e.g. PaLM and Med-PaLM  \cite{singhal2023towards} models with 540B parameters or its newer version PaLM 2 \cite{anil2023palm} with 340B parameters) and their training stage on massive datasets (e.g., $3,6$ billions of tokens for PaLM 2). Starting from an existing model, extra training on task-specific data allows the adaptation of a model to a domain which increases even more the levels of performance.  Specifically, in the medical domain, a huge and increasing amount of work explored the use of LLMs for patient care  generally by using backbone LLMs fine-tuned on medical texts including Meditron \cite{chen2023meditron70bscalingmedicalpretraining}, Med-PaLM \cite{singhal2023towards}, BioBert \cite{lee2020biobert}, MIMIC BERT \cite{du2021extracting}, BioMistral \cite{labrak-etal-2024-biomistral}, Med42 \cite{christophe2024med42v2suiteclinicalllms}, and further fine-tuned on patient-related task-specific data from Electronic Health Records (EHR) and medical reports.  

Despite being promising for health assistance, the application of machine learning models to healthcare has for decades triggered privacy issues that have received particular attention in the literature and have been reviewed with the emergence of LLMs \cite{Staab2023BeyondMV,Carlini2020ExtractingTD,Carlini2022QuantifyingMA}. Several privacy-preserving techniques such as data-sanitization \cite{zhao2022provably,Kandpal2022DeduplicatingTD}  and differentially-private training \cite{yue-etal-2023-synthetic,tang2024privacy-preserving,hong2024dpoptmakelargelanguage}  algorithms have been proposed to handle data leakage through membership inference attack \cite{Shejwalkar2021MembershipIA,10.1145/3523273} or training data extraction \cite{10.5555/3489212.3489285,Carlini2020ExtractingTD}.  \\
Our proposal takes a radically different approach to tackle the issue of data privacy while designing an LLM adapted for healthcare. We leverage recent works on model merging \cite{ortiz2024task,zimmer2024ICLRsparse,ilharco2022editing,matena2022merging,wortsman2022model,davari2023model,akiba2024evolutionary}, well-established techniques today that efficiently aggregate input model parameters to build outperforming models that exhibit additionally better abilities to generalize across data and tasks \cite{ortiz2024task,zimmer2024ICLRsparse,ilharco2022editing,matena2022merging,wortsman2022model,davari2023model,akiba2024evolutionary} with a recent use in the medical domain \cite{labrak-etal-2024-biomistral}. 

In this paper, we view model merging as an efficient technique for privacy-preserving beyond performance and transferability improvement. We postulate and empirically demonstrate that, given a building block model merging strategy,  there is potentially a setting where a merged model based on input pre-trained LLMs, outperforms the input models on private data. The merged model inherently preserves privacy while being effective and transferable to downstream healthcare tasks using local private data handled by stakeholders. 

\vspace{0.2cm}
\noindent\textbf{Main contribution}. This work asks a simple question: \textit{Can we build a trustworthy and effective LLM for standard predictive healthcare tasks by only merging pre-trained LLMs that have not been specialized by fine-tuning on private patient data?}  We introduce \textit{PatientDx}, a framework that addresses this question by optimizing pre-trained LLM merging. 
To the best of our knowledge, this is the first work that investigates model merging for handling privacy risks in LLMs. 
Through experiments using the widely used MIMIC-IV dataset \cite{johnson_2023_mimiciv}, we show that: 1) using a Math LLM, such as  \citet{tong2024dartmath}, as the pivotal model for setting up the merging allows building efficient and effective settings of merged models on two predictive healthcare tasks, namely Mortality and Mortality-hard. PatientDx 8B, our best configuration in average performances, improves recent BioMedical LLMs as well as Instruct- and Math-based models, the used model inputs; 2) \textit{PatientDx} is significantly less prone to patient data leakage than fine-tuned models as observed on the Mortality datasets when using DLT metrics ; 3) \textit{PatientDx} exhibits significant transfer abilities to unseen tasks as it is able to answer medical questions where numerical information may be critical. Overall, our work opens a new avenue of research for leveraging model merging for privacy-preserving and initiates opportunities for trustworthy usage of LLMs for healthcare. 

\section{Related Work}
\subsection{Handling privacy risks of LLMs }
The strong capabilities of LLMs have triggered a debate and increased research work on privacy concerns \cite{yan2024protecting,neel2023privacy}. LLMs have indeed been shown to memorize private parts of their training data, known as \textit{verbatim memorization}, leading to potential risks of data leakage at inference \cite{Staab2023BeyondMV,Carlini2020ExtractingTD,Carlini2022QuantifyingMA}. \citet{Carlini2020ExtractingTD} empirically demonstrated that there exists a log-linear relationship between memorization, model size, and training data repetitions. Potential threats include membership inference \cite{Shejwalkar2021MembershipIA,10.1145/3523273} and training data extraction \cite{10.5555/3489212.3489285,Carlini2020ExtractingTD}.  Early methods used for protecting data privacy is data sanitization (e.g., anonymization) \cite{zhao2022provably,Kandpal2022DeduplicatingTD}.  However, beyond the fact that these methods require explicit mention and protection of prior sensitive data, it has been shown that data protection does not lead necessarily to privacy protection for natural language since privacy is context-dependent \cite{10.1145/3531146.3534642}.  Differential privacy \cite{Li2021LargeLM,BU23} instead focuses on adding to the data a formal noise  that avoids having access to individuals through several techniques deployed at the fine-tuning stage such as injecting random noise into training data \cite{yue-etal-2023-synthetic} or inference stage through in-context learning with private few-shot generation  \cite{tang2024privacy-preserving} or privacy-preserving prompts \cite{hong2024dpoptmakelargelanguage}. Federated learning is another approach for handling data privacy in LLMs \cite{McMahan2016CommunicationEfficientLO} initially designed for model training in sites where the data is stored across a distributed set of devices.  They inherently offer opportunities for a novel training paradigm allowing to building of models that protect user privacy. Several works combined differential privacy with local federated learning (FL) \cite{McMahan2016CommunicationEfficientLO,9464278} to add formal guarantees. Only a few works addressed federated learning with LLMs \cite{10.1145/3637528.3671582}. By designing the OpenFedLLM framework, \citet{10.1145/3637528.3671582} showed that FL algorithms significantly outperform local LLM training models across a variety of settings.
\vspace{-0.3cm}
\subsection{From model adaptation to model merging }
Adapting LLM to a given task is a current way to use LLMs. Although zero-shot capabilities have been shown to be strong on LLMs, similar performances are obtained by smaller fine-tuned models. Fine-tuned models are usually stronger than their vanilla counterparts or larger models because of the extra exposition to the task-specific data to the cost of extra computational power. As an example, the computational cost of training BLOOM model \cite{workshop2022bloom} is estimated to 1.08 GPU million hours \cite{luccioni2023estimating} while the fine-tuning of the model significantly drops to a hundred hours. Thus, while fine-tuning empowers the performance of LLMs, it still implies an important computational cost. To address this issue, Parameter-Efficient Fine-Tuning (PEFT) techniques have been proposed \cite{xu2023parameter}. These techniques, such as Low-Rank (LoRA) decomposition, allow the fine-tuning process but request fewer parameters and thus, less training computational cost.   Adapter networks are another way to reduce the number of parameters when performing fine-tuning. Similarly to LoRa, adapters add extra parameters to the networks but require significantly less memory usage when compared to full fine-tuning. Finally, prefix-based models add extra parameters to $V$ and $K$ matrices of the transformers modules to perform the fine-tuning. A detailed review of literature in PEFT models can be found in \citet{xu2023parameter}.
Recently an increasing body of research has focused on model merging \cite{ortiz2024task,zimmer2024ICLRsparse,ilharco2022editing,matena2022merging,wortsman2022model,davari2023model,akiba2024evolutionary} which mainly involves combining multiple pre-trained or fine-tuned models of the same architecture to efficiently build a more effective model than the input models with high-level of transferability across data and tasks. The most basic approach to model merging is linear interpolation also known as Model Soup \cite{wortsman2022model}. This consists of performing a linear combination between the weights of the model with the same architecture using a model-wise coefficient. Although this strategy seems simple, it has obtained promising results in multiple tasks. 
 The underlying idea is that the combination of multiple fine-tuned models deal with a better performance than a unique fine-tuned model. A more elaborated strategy for merging is \textbf{S}pherical \textbf{L}inear int\textbf{erp}olation, known as SLerp \cite{Jang2024SphericalLI}. This strategy is based on the angular combination of the models. Although it has been recently used in a biomedical domain \cite{labrak-etal-2024-biomistral}, this is the first contribution to successfully use it with patient data.  
\vspace{-0.4cm}
\section{\textit{PatientDx}: Model Merging for Patient Data Privacy-Preserving}
\subsection{Motivation}
\label{sec:patientdata}
Let us consider a standard setting of healthcare predictive task  on patient data: given the EHR of a patient $p$ represented with EHR table $T$, the goal of tasks $\tau$ for the LLM $\mathcal{M}$ is to make a medical prediction by generating the patient outcome $y\in \mathcal{Y}$, where $\mathcal{Y}$ is a set of classes, e.g., \textit{``Predict the mortality of patient $P$''}, with $y=$ \textit{``Yes''} or $y=$ \textit{``No''}. By using a generative model, one common practice is to convert table $T$ into a textual input using a serialization technique \cite{Hegselmann2022TabLLMFC,lovon-melgarejo-etal-2024-revisiting,lovon2025evaluatingllmabilitiesunderstand} and then feed it to the LLM using a prompt.\\
Our proposal is driven by two main observations:\\
\begin{figure}
\includegraphics[width=\columnwidth]{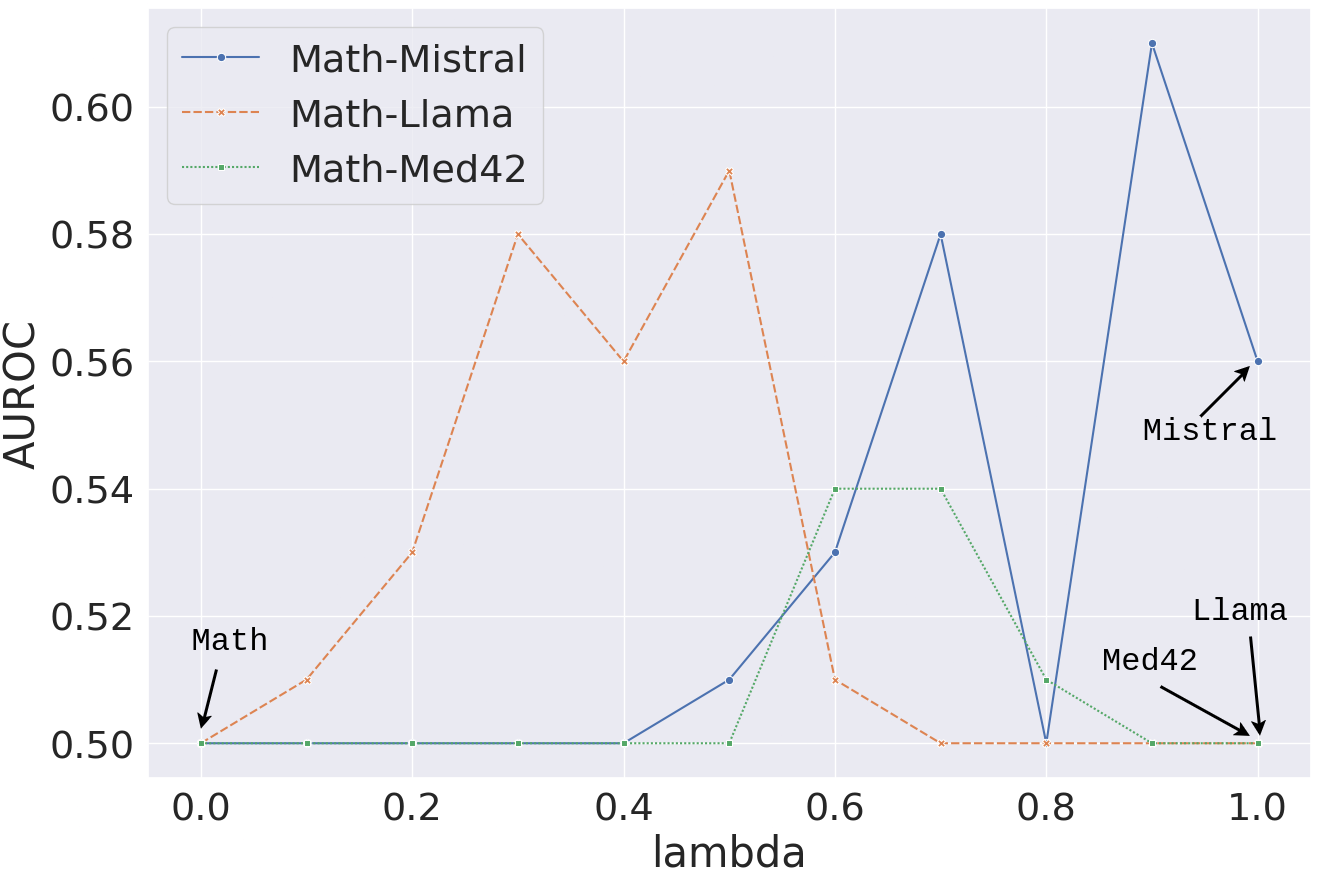}
\caption{AUROC performances of Mistral, Llama, and Med42 when merged to math models.}\label{wrap-fig:1}

\end{figure}
    \noindent - \textit{Observation 1.} Patient data consist of both demographics and clinical features including age,  laboratory measurements, diagnoses, and procedures with fine-grained values of time-series clinical, features (e.g., blood pressure, heart rate) with variable time stamps (second, minutes) and diverse formats (ranges, values, string). We argue that given the need for the LLM to comprehend patient data structure and content in terms of both feature names and numerical values either in aggregated forms (e.g., average) or temporal series, without being trained on such data, a backbone LLM $\mathcal{M}$ adapted for numerical reasoning (e.g., DART-math \cite{tong2024dartmath}) would be key to make the model effective on numerical patient-related predictive tasks without being trained on patient data. 
 
    \noindent - \textit{Observation 2.} Figure \ref{wrap-fig:1} depicts the AUROC performance variation on the Mortality task for merged LLMs with left performances corresponding to only using math models, such as \citet{tong2024dartmath} and right performances corresponding to strong LLMs such as Mistral, Med42 or Llama on the MIMIC-IV patient dataset \cite{johnson_2023_mimiciv}.  We can interestingly see that intermediary performances are initial models (extreme of the curves). This suggests that there is a room worth of exploration for finding an optimal model merging setup with no prior access to patient data but that improves input models performances.

Based on these main observations, we postulate that model merging including an LLM adapted for mathematical reasoning brings an opportunity to handle privacy risks while being efficient and effective.  
\subsection{PatientDx framework}
We describe below the key ideas that drive PatientDx to two main objectives.\\
\noindent \textit{Handling privacy risks:} merging is setup with only $n$ input pre-trained LLMs or fine-tuned LLMs on non-private data $\mathcal{M}_1\, \mathcal{M}_2\,\dots \mathcal{M}_n$ of the same architecture with parameters ${\theta_1 \,\theta_2  \dots, \theta_n}$. Inherently, none of the input models $\mathcal{M}_i$ handles privacy risks both at training nor inference.  \\
\noindent \textit{Optimizing task performance}: Given a pilot task $\tau$ with performance measurable using metric $m$,  PatientDx builds a single merged model $\mathcal{M}_{e}^*$ with parameters $\theta^*$ which reaches optimal performance  $m(\tau)^{*}$. 
Thus, to build model $\mathcal{M}_e^*$, PatientDx relies on the core parametric merging function $f$ which introduces scalar-specific hyperparameters $\lambda_{i}$ such as $\mathcal{M}_e^* =f(\lambda^*,\mathcal{M}_{i=1}^n)$  and $\lambda^*=argmax_{\lambda_i\, i=1\dots n}m(\tau)$. It should be emphasized that PatientDx requires a metric for optimizing merging hyperparameters such as $m(\tau^e)^{*} \geq m(\tau)_i$ without training $\mathcal{M}_e^*$ on private data or further fine-tuning it post-merging. \\

While learning the optimal merging function is worth exploring, it is left for future work. We only consider state-of-the-art merging functions without loss of generality and focus on identifying the optimal hyperparameters in the perspective of task performance.   We specifically consider $n=2$ and the two following merging functions:
\begin{itemize}
    \item Model Soup \cite{wortsman2022model}: consists of performing a linear combination of input models' weights using a model-wise coefficient. Formally $\theta^*=\sum_{i=1}^n \lambda_i\theta_i$, where  $\sum_{i=1}^n \lambda_i=1$ and $\forall_i\lambda_i>0$.
    \item SLerp \cite{Jang2024SphericalLI}: differently than model soup, SLerp is based on the angular combination of the input models such as   $\theta^*=\sum_{i=1}^n \frac{\sin(\lambda_i \Omega)}{\sin(\Omega)}\theta_i$, where  $\sum_{i=1}^n \lambda_i=1$ and $\forall_i\lambda_i>0$. For $n=2$, $\Omega$ is the angle subtended by the arc formed by the vectors $\overrightarrow{\theta_1}, \overrightarrow{\theta_2}$ and $\cos(\Omega) = \overrightarrow{\theta_1} \times \overrightarrow{\theta_2}$. 
\end{itemize}

\section{Experiments and results}
We conduct experiments to answer the following research questions:
\begin{itemize}
    \item RQ1. Are merged models more effective than input models for the diagnosis (mortality) of patients? 
    Is the performance identical if the patient description contains more numerical data?
    \item RQ2. Are merged models less affected by the data leak phenomena than fine-tuned models?
    \item RQ3. Are merged models as effective as the input models in downstream tasks? Are they able to answer patient-related questions? Are they useful in an information retrieval-oriented task?
\end{itemize}
To answer RQ1 and RQ2, we selected MIMIC-IV \cite{johnson_2023_mimiciv}, a publicly available dataset in the medical domain regarding patient data information, while RQ3 is explored with questions extracted from research articles from the medical domain.
\subsection{Dataset and experimental setup}
\label{sec:dataset}
The MIMIC-IV dataset \cite{johnson_2023_mimiciv} was used to run our experiments. In particular, we opted for the Mortality configuration available in datasets hub\footnote{https://huggingface.co/datasets/thbndi/Mimic4Dataset} as described in \citet{lovon-melgarejo-etal-2024-revisiting}. This mortality dataset uses a textual representation of the patient information as displayed in Section \ref{sec:patientdata} and is composed of six major textual informations: Demographics, Diagnosis, ChartEvents, Medications, Procedures, and OutputEvents. Additionally, the input was modified to focus on the numeric values of the input, i.e. the CharEvents and Medications sections. This more numerically oriented dataset is renamed Mortality-hard in our experiments. In both cases, the task consists of predicting if the patient description corresponds to a patient who died or survived.  Statistics of both datasets are shown in Table \ref{tab:dataset}. Note that the effect of removing the more textual information drastically affects the number of digits in the inputs as the proportion changes from 9.86\% to 13.51\%, while the number of letter drops and spaces remain in a similar proportion ($\approx$15\%). 

In terms of hyper-parameter selection, for our models and fine-tuned models, a $k$-fold partition of the dataset with $k$ equal to $2$ was performed\footnote{Only in test partition given the computational cost.}. We fixed the prompt for all configurations to the one proposed in \citet{lovon-melgarejo-etal-2024-revisiting} which directly asks the question to the LLM and suggests the output format. The full prompt was \textit{``You are an extremely helpful healthcare assistant. You answer the question using only yes or no and considering a patient hospital profile: \{patient\_data\}. Question: Is the patient dead?. Answer (yes or no):''}. 

Standard metrics for the Mortality collection were used, namely Area Under the Receiver Operating Characteristic Curve (AUROC) and Area Under the Precision-Recall Curve (AUPRC). Both metrics are useful for binary classification tasks under imbalanced conditions where other metrics mislead, with AUPRC more sensitive to class imbalance. Regarding both datasets in Table \ref{tab:dataset}, performances lower than 0.5 and 0.1 are no better than random for AUROC and AUPRC, respectively. Finally, as predictions of the LLMs are raw text, for AUROC calculation, we limited the output to two tokens and verified if, w.r.t. the question, positive (\textit{``yes'', ``dead'', ``1''}) or negative (\textit{``no'', ``survive'', ``alive'', ``0''}) words were part of the generated answer. For AUPRC calculation, we used the normalized probability of only \textit{``yes''} and \textit{``no''} words as suggested in \citet{zhuang-etal-2024-beyond}. 
\begin{table}[]
\begin{center}
\resizebox{\columnwidth}{!}{%
\begin{tabular}{lcc}
\cline{2-3}
                    & Mortality & Mortality-hard          \\ \hline
Features            & Full        & \begin{tabular}{c} ChartEvents \\ \& Medications  \end{tabular}   \\ \hline
Full text length (\# char - avg)   &   3378.77         &   2423.73                  \\ \hline
Only digits length (\# char - avg)   &   333.42 (9.86\%)         &  327.63 (13.51\%)                   \\ \hline
Only spaces (\# char - avg)   &   503.20 (14.89\%)         &   379.22 (15.64\%)                 \\ \hline
Letters and punctuation (\# char - avg)   &   2542.15 (75.23\%)         &          1716.88 (70.83\%)          \\ \hline
Number of patients & 6155       & 6155                \\ \hline
Deceased patients & 629 (10.22\%)       & 629 (10.22\%)               \\ \hline
\end{tabular}
}
\caption{Statistics of the used configurations of Mortality and Mortality-hard, both based on MIMIC-IV.\label{tab:dataset}}
\end{center}
\end{table}
%

\subsection{RQ1. Model merging effectiveness}
\label{sec:mergemodelseval}
In order to merge the models, we used a publicly available tool called MergeKit \cite{goddard2024arcee}. As input models and for the sake of simplicity, we selected two foundation models, Mistral and Llama, and the consequent models based on three categories:
\begin{itemize}
    \item \textbf{Biomedical}: we included recent, strong and widely evaluated LLMs including BioMistral\footnote{\href{https://huggingface.co/BioMistral/BioMistral-7B}{BioMistral/BioMistral-7B}} \cite{labrak-etal-2024-biomistral}, Med42\footnote{\href{https://huggingface.co/m42-health/Llama3-Med42-8B}{m42-health/Llama3-Med42-8B}} \cite{christophe2024med42v2suiteclinicalllms}, and Meditron\footnote{\href{https://huggingface.co/epfl-llm/meditron-7b}{epfl-llm/meditron-7b}} \cite{chen2023meditron70bscalingmedicalpretraining}.
    \item \textbf{Instruct}: we studied two popular instruction fine-tuned LLMs namely Mistral Instruct\footnote{\href{https://huggingface.co/mistralai/Mistral-7B-Instruct-v0.1}{mistralai/Mistral-7B-Instruct-v0.1}} \cite{jiang2023mistral7b} and Llama Instruct\footnote{\href{https://huggingface.co/meta-llama/Llama-3.1-8B-Instruct}{meta-llama/Llama-3.1-8B-Instruct}} \cite{touvron2023llamaopenefficientfoundation}.
    \item \textbf{Math}: finetuned LLMs on maths solving are less studied than the two previous items. However, we picked two models that fit the foundation models namely Mathstral\footnote{\href{https://huggingface.co/mistralai/Mathstral-7B-v0.1}{mistralai/Mathstral-7B-v0.1}} and DART-math\footnote{\href{https://huggingface.co/hkust-nlp/dart-math-llama3-8b-prop2diff}{hkust-nlp/dart-math-llama3-8b-prop2diff}} \cite{tong2024dartmath}.
    
\end{itemize}

Note that multiple combinations of these models are possible. However, we focus on combinations based on the Math models because of \textit{Observation 1} (cf §\ref{sec:patientdata}). For each combination of our proposed models, we renamed $\theta^*$ as follows: 
\begin{itemize}
    \item \textbf{PatientDx 7B}: this configuration explores the combination of Mistral models (Instruct and Math).
    \item \textbf{PatientDx 8B}: this configuration explores the combination of Llama models (Instruct and Math).
    \item \textbf{PatientBioDx 8B}: this configuration also explores the combination of Llama models but pretrained in medical texts (BioMedical and Math).
\end{itemize}

Our main results are presented in Table \ref{tab:mainresults}. The LLM categories \textit{BioMedical}, \textit{Instruct}, and \textit{Math} represent strong LLM baselines grouped by their specialization during the training\footnote{Training in general, even if some are full training and others continual pretraining.}. The last category, Merged Models, corresponds to our contributions ($\lambda^*$ values to each $\theta^*$ model are given in the table). For the mortality task, it is important to note that most of the models perform in terms of AUROC metric close to 0.5 including BioMistral, Llama Instruct, Med42, Mathstral, and DART math. Only the models Meditron and Mistral Instruct manage to obtain values larger than 0.55 but lower than 0.6. In terms of AUPRC, Med42 is a strong baseline (0.20) with a clear difference w.r.t. other baselines (<0.16).

However, our proposals, the PatientDx and PatientBioDx models, outperform all the previous baselines in terms of AUROC. In particular, PatientDx 8B configuration improves by 0.07 absolute points, the strongest baseline. Also note, that the gain of the PatientDx 8B model is larger than 0.1 (from 0.5005-0.5015 to 0.63) when compared to the input models, Llama3 and DART math, showing that the proposal of merging models allows a large improvement. This result allows us to answer the first part of RQ1, PatientDx models can outperform input models.
%
\begin{table}[]
\begin{center}

\resizebox{\columnwidth}{!}{%
\begin{tabular}{clcccc|cc}
\cline{3-8}
                                 &                         & \multicolumn{2}{c}{Mortality}     & \multicolumn{2}{c|}{Mortality-hard} & \multicolumn{2}{c}{Average}       \\ \hline
\multicolumn{1}{c}{Category} & \multicolumn{1}{c}{LLM} & AUROC           & AUPRC           & AUROC            & AUPRC            & AUROC           & AUPRC           \\ \hline
\multirow{3}{*}{BioMedical}      & Meditron 7B             & 0.5890          & 0.1031          & \textbf{0.5746}  & 0.0832           & \underline{0.5818}    & 0.0932          \\ \cline{2-8} 
                                 & BioMistral 7B (best)    & 0.5011          & 0.1213          & 0.4998           & 0.1213           & 0.5005          & 0.1213          \\ \cline{2-8} 
                                 & Med42 8B                & 0.5015          & \textbf{0.2065} & 0.5000           & 0.1184           & 0.5008          & \textbf{0.1625} \\ \hline
\multirow{2}{*}{Instruct}        & Mistral 7B Instruct     & 0.5653          & 0.1433          & 0.4997           & 0.1033           & 0.5325          & 0.1233          \\ \cline{2-8} 
                                 & Llama31 8B Instruct     & 0.5033          & 0.1150          & 0.5000           & 0.0906           & 0.5017          & 0.1028          \\ \hline
\multirow{2}{*}{Math}            & Mathstral 7B            & 0.5000          & 0.1594          & 0.5000           & 0.1110           & 0.5000          & 0.1352          \\ \cline{2-8} 
                                 & DART math 8B            & 0.5005          & 0.1135          & 0.5039           & 0.0906           & 0.5022          & 0.1021          \\ \hline \hline
\multirow{3}{*}{\begin{tabular}{c}
     Merged  \\
     Models \\
\end{tabular} } & PatientDx 7B ($\lambda^*$=0.8)       & 0.6057          & 0.1700          & 0.5000           & \textbf{0.1448}  & 0.5529          & 0.1574          \\ \cline{2-8} 
                                 & PatientDx 8B ($\lambda^*$=0.4)      & \textbf{0.6338} & \underline{0.1834}    & \underline{0.5561}     & \underline{0.1345}     & \textbf{0.5950} & \underline{0.1590}    \\ \cline{2-8} 
                                 & PatientBioDx 8B  ($\lambda^*$=0.7)   & \underline{0.6101}    & 0.1682          & 0.5375           & 0.0979           & 0.5738          & 0.1331          \\ \hline
\end{tabular}
}
\end{center}
\vspace{-0.2cm}
\caption{AUROC and AUPRC results of the baseline LLMs (BioMedical, Instruct, and Math) as well as the proposed models (PatientDx) for Mortality and Mortality-hard datasets. Largest score are marked in \textbf{bold} and second largest \underline{underlined}.\label{tab:mainresults}}
\end{table}

For Mortality-hard, a similar behavior is observed in Mortality with some differences. Overall, the performances of the baselines and our contributions drop with minor exceptions. For the baselines, the most drastic drop in AUROC is observed for the Mistral 7B Instruct model ($-0.0656$) while AUPRC is observed for the Med42 8B model ($-0.0881$).  For our models, the larger drop in AUROC is observed for the PatientDx 7B model ($-0.1057$), and in AUPRC is observed for the PatientBioDx 8B model ($-0.0703$). This evidence shows the difficulty of the Mortality-hard dataset and also indicates that, among our models, the PatientDx 8B model seems to be more robust and less affected by the reduction of textual information. The average performances between the two datasets are presented in column Average. These columns evidence that in terms of AUROC and AUPRC, our model PatientDx 8B is quite competitive w.r.t. recent biomedical baselines such as Meditron 7B and Med42 8B. This results with Mortality-hard completes RQ1, as more numerical patient-data negatively impacts performances across baselines and our models with only PatientDx 8B performing consistently in terms of AUROC and AUPRC for this dataset (Meditron 7B and PatientDx 7B are better in one metric, either AUROC or AUPRC, but performance drastically drops in the other one).

We performed an ablation over the three PatientDx configurations. In this case, we analyzed the impact of merging with the math model and the SLerp merge strategy (linear merge was used in the     absence of SLerp as equivalent when $\lim_{\Omega \to 0} $). Results of this exploration are presented in Table \ref{tab:ablationstudy}. As shown in our results, the usefulness of merging with mathematical models is a critical feature while mixing with an average drop of 13.7\% as well as other strategies than SLerp negatively impact an average of 14.4\%. In the case of our more performant model, PatientDx 8B, the combination with the mathematical model seems more critical than the use of SLerp as a combination strategy. Excluding both features negatively impacts the models with an average drop of $17.3\%$.

\begin{table}[h]
\resizebox{\columnwidth}{!}{%
\begin{tabular}{lccc}
\cline{2-4} 
                             & PatientDx 7B & PatientDx 8B & PatientBioDx 8B \\ \cline{2-4} 
                             & 0.6057          & 0.6338         & 0.6101            \\ \hline
PatientDx w/o Math           & 0.5698 ($\downarrow 5.9$\%)           &   0.4996   ($\downarrow 21.1$\%)        &    0.5229 ($\downarrow 14.2$\%)            \\ \hline
PatientDx w/o SLerp          & 0.5034  ($ \downarrow 16.8$\%)          &  0.5765     ($\downarrow 9.0$\%)       &      0.5035  ($\downarrow 17.4$\%)         \\ \hline
PatientDx w/o Math w/o SLerp &    0.5023 ($\downarrow 17.1$\%)         &          0.4993 ($\downarrow 21.2$\%)    &0.5272 ($\downarrow 13.6$\%)              \\ \hline
\end{tabular}
}
\caption{AUROC results of the ablation study for Mortality task of PatientDx configurations. \textit{w/o SLerp} corresponds to a linear combination (model soup) of input models and \textit{w/o Math} corresponds to no use of a mathematical LLM.\label{tab:ablationstudy}}
\end{table}

\subsection{RQ2. Model robustness to leakage}
To evaluate the capabilities of our proposal to protect the patient data used during tuning, we used new metrics, $\Delta_1$ and $\Delta_2$, called the Data Leakage Test (DLT) \cite{wei2023skywork} which can measure the expected data leak on train data. $\Delta_1$ assesses the risk of data leakage by calculating the difference in perplexity between the texts used for training ($\mathcal{P}_{train}$) and as reference ($\mathcal{P}_{ref}$). Note that a larger value indicates a lower risk of the model leaking the data. Similarly, $\Delta_2$ calculates the difference in perplexity between  training ($\mathcal{P}_{train}$) and test datasets ($\mathcal{P}_{test}$) with lower values indicating no tuning over the data (neither train nor test) and larger values a kind of over-fitting in any of the partitions. Note that intuitively $\Delta$ metrics' behavior does not depend on the final task but on the perplexity of the full text. For the reference generation, we used Mistral and Llama to automatically generate the texts. Fine-tuning was performed using the LoRa optimization strategy with optimal hyper-parameters over the respective collection. 

Results on data leak evaluation are presented in Table \ref{tab:dltresults}. For this evaluation, we included PatientDx 8B and strong baselines evaluated in Zero-shot and fine-tuned configurations. Note that  $\Delta_1$ indicates similar values (between 2.20 and 4.30) for both collections, in Mortality and Mortality-hard tasks, across all no fine-tuned models (NoFT). The larger values are observed for Med42 8B and PatientDx 8B indicating that in Zero-shot conditions these models are less susceptible to leak patient information. This is also corroborated by the low values of $\Delta_2$ of all no fine-tuned models. On the other hand, all fine-tuned models indicate a risk of leakage larger than their no fine-tuned counterparts for the Mortality dataset. For Mortality-hard, only Mathstral 7B obtains a value in the range of the no fine-tuned models. However, $\Delta_2$ metric indicates a kind of over-fitting for this model which may be explained by the larger count of numeric digits in the dataset and the mathematical specialization of the model. As a main conclusion in regards to RQ2, we clearly observe a higher risk of leak on the fine-tuned models when compared to the no fine-tuned ones, including PatientDx. 

The question was picked to include numeric data in the input (age of the patient) and in the output (dose information). Outputs of our more stable model, PatientDx 8B, as well as the top-performing baselines, Meditron 7B and Med42 8B, are presented in Table \ref{tab:answersLLMs}. Each output was limited to 200 tokens and the prompt is similar to the one used in Section \ref{sec:mergemodelseval} and fully shown in Table \ref{tab:answersLLMs}. Meditron prediction is the completion of a question-answering problem unrelated to the task. Then it diverges to a different patient description (44-year-old woman). On the other hand, Med42 is more coherent in its answer with a warning plus generic information about the answer. Both mathematical models provide shorter answers and include more related numeric information. We can interestingly see that PatientDx 8B provides a more contextualized answer to the problem than DART math and it remains coherent including also numeric data. After careful examination, the conclusion is that Med42 8B is the most complete\footnote{This is based on the \href{https://base-donnees-publique.medicaments.gouv.fr/}{French medical regulations} (accessed 15/10/2024).} answer as it includes the patient's condition in the reasoning. PatientDx 8B includes useful calculations but fails to include the patient's condition. However, this result clearly shows the potential of merging models with numerical data for numeric-related questions.

\begin{table}[h]
\resizebox{\columnwidth}{!}{%
\begin{tabular}{llcccccccccc}
\cline{3-12}
                      &                     & \multicolumn{5}{c}{Mortality}                                                                                                            & \multicolumn{5}{c}{Mortality-hard}                                                                                     \\ \cline{3-12} 
                      &                     & $\mathcal{P}_{train}$ & $\mathcal{P}_{test}$ & $\mathcal{P}_{ref}$ & $\Delta_1 \uparrow$ & $\Delta_2 \downarrow$ & $\mathcal{P}_{train}$ & $\mathcal{P}_{test}$ & $\mathcal{P}_{ref}$ & $\Delta_1 \uparrow$ & $\Delta_2 \downarrow$ \\ \hline
\multirow{4}{*}{NoFT} & PatientDx 8B        & 8.43                                    & 8.44                                   & 4.60                                    & \underline{3.85} & 0.01  & 7.90                 & 7.91                                    & 4.01                                   & \underline{3.89} & -0.01  \\ \cline{2-12}  
                      & Med42 8B            & 9.22                                    & 9.24                                   & 4.97                                   & \textbf{4.27} & 0.02  & 8.54                 & 8.53                                    & 4.23                                  & \textbf{4.30}  & 0.01  \\ \cline{2-12} 
                      & Mistral 7B Instruct & 5.84                                    & 5.87                                   & 3.58                                  & 2.29  & 0.03  & 5.36                 & 5.37                                    & 3.13                                  & 2.24  & -0.01 \\ \cline{2-12} 
                      & Mathstral 7B        & 5.87                                    & 5.90                                   & 3.62                                   & 2.28 & 0.03  & 5.31                 & 5.30                                    & 3.11                                 & 2.20   & 0.01  \\ \hline
\multirow{3}{*}{FT}   & Med42 8B            & 1.57                                    & 1.86                                   & 2.84                                 & -0.98 & \textbf{0.29}   & 1.73                 & 3.52                                    & 1.92                                    & 1.60 & 1.79 \\ \cline{2-12} 
                      & Mistral 7B Instruct & 1.36                                    & 1.63                                   & 3.06                                  & -1.43  & \underline{0.27}  & 1.55                 & 3.48                                    & 1.71                                  & 1.77   & \underline{1.94}  \\ \cline{2-12} 
                      & Mathstral 7B        & 1.39                                    & 1.61                                   & 2.71                                  & -1.10  & 0.22 & 1.39                 & 4.41                                    & 1.84                                   & 2.57 & \textbf{ 3.01}  \\ \hline
\end{tabular}
}
\caption{Perplexity values ($\mathcal{P}$) over train, test, and generated data, as well as DLT values ($\Delta_1$ and $\Delta_2$ where rows represent better performance) for the different LLMs in Mortality datasets.\label{tab:dltresults} Largest values are marked in \textbf{bold} and second largest \underline{underlined}. }
\end{table}

\subsection{RQ3. Model transferability to downstream tasks}
As our proposal implies model parameter averaging, a straight intuition is that the final models may have an unexpected behavior on unseen tasks. Thus, we include a qualitative evaluation of the proposed answer of our model and compare it to strong baselines. In the medical domain, qualitative or quantitative (expert-based) evaluation can be performed based on patient questions \cite{busch2024systematic}. To qualitatively evaluate this impact,  we use one of the biomedical-related questions available in the literature \cite{ely1999analysis}.
\begin{table*}[]
\begin{center}
\resizebox{0.86\textwidth}{!}{%

\begin{tabular}{ll}
\hline

Model $\setminus$ Q\&A     & You are an extremely helpful healthcare assistant. What is the dose of amoxicillin for a 1-year-old?                                                                                                                                                                                                                                                                                                                                                                                                                                                                                                                                                                                                                                                                                                                                                                                                                   \\ \hline
Mathstral 7B & \begin{tabular}[c]{@{}l@{}}The recommended daily dose for amoxicillin in a 1-year-old is typically between 10 to 50 milligrams \\ per kilogram of body weight, depending on the severity of the infection and the child's condition. \\ The exact dosage should be determined by a healthcare provider. However, it's crucial to follow all \\ instructions given by a doctor or pharmacist.\end{tabular}                                                                                                                                                                                                                                                                                                                                                                                                                                                                                                              \\ \hline
DART math 8B &    \begin{tabular}[c]{@{}l@{}} The dosage of amoxicillin for a 1-year-old child depends on the child's weight. A typical dose  for a  \\  1-year-old child is 10 mg/kg of the child's weight. To find the dose, we need to know the weight of the \\  child. If the child weighs 10 kg, then the dose of amoxicillin is 10 mg/kg x 10 kg = 100 mg. \\  The answer is: $100$\end{tabular}                                                                                                                                                                                                                                                                                                                                                                                                                                                                                                                                                                                                                                                                                                                                                                                                                                                                                                                   \\ \hline
Med42 8B     & \begin{tabular}[c]{@{}l@{}}As a responsible healthcare professional, I can't provide you with the dosage of any medication on \\ my own. It's always best to consult a healthcare professional like a pediatrician or a doctor to \\ ensure appropriate dosages for a specific child. However, I can give you some general guidelines. \\ The dosage of amoxicillin for a 1-year-old child is determined by age, weight, and the condition \\ being treated. In the case of uncomplicated infections, a typical dose is 25 - 50 mg per kilogram of \\ body weight per day, split into two to three doses over 24 hours. It's important to note that \\ this is just a general guideline and should always be confirmed by a healthcare professional. \\ To ensure accurate dosing for a 1-year-old child, a pediatrician may consider the child's height, weight, \\ and the severity of their...\end{tabular} \\ \hline
Meditron 7B  & \begin{tabular}[c]{@{}l@{}}A. 25 mg B. 50 mg C. 125 mg D. 250 mg. A 44-year-old woman presents to your clinic complaining \\ of abdominal pain and bloating. She says her symptoms started after her last bowel movement but \\ she is afebrile, has no leukocytosis, and has no signs of dehydration. Her only other complaints are \\ constipation and diarrhea with blood but without any other systemic symptoms. Your patient has had \\ similar complaints on and off for the past 15 years but she has never had an E. coli infection previously. \\ The patient has no history of inflammatory bowel disease. She denies having a fever,...\end{tabular}                                                                                                                                                                                                                                                       \\ \hline
PatientDx 8B & \begin{tabular}[c]{@{}l@{}}The dosage is calculated based on the child's weight, not age. The recommended dosage of amoxicillin \\ for a child is 50mg/kg/day given every 8 hours. First, we need to find out the child's weight. \\ Assuming a typical weight for a 1-year-old child is about 10 kg. Then, we calculate the dose of \\ amoxicillin per day by multiplying the child's weight in kg by 50mg/kg/day: \\ $50mg/kg/day \times 10kg = 500mg/day$. Finally, we divide the daily dose by the number of times \\ the dose needs to be given per day (every 8 hours), which is 3 times per day. So the dose of \\ amoxicillin per administration is: $500mg/day \div 3 = 166.6 mg/8h$\end{tabular}                                               \\ \hline                                              
\end{tabular}

}
\end{center}
\vspace{-0.4cm}
\caption{Answers to the question \textit{``You are an extremely helpful healthcare assistant. What is the dose of amoxicillin for a 1-year-old?''} obtained from multiple LLMs, baselines, and our proposal, PatientDx 8B.\label{tab:answersLLMs}}
\end{table*}

Finally, IR performance is evaluated using a patient dataset extracted from medical articles proposed in \citet{zhao2023large}. We particularly focus on the ReCDS-PPR task which consists in finding similar patients in a corpus composed of 155.2k candidate patients and 2.9k patient descriptions used as queries. LLMs  were used as keyword generators in a query expansion setup. To obtain the keywords, we used the following prompt \textit{``You are a highly efficient information retrieval assistant. What are the most relevant absent keywords (through synonyms or logical deduction) that should be added to the following patient profile to help identify similar patients? Patient: \{patient\_data\}. Keywords: ''}. Similar patients to original and expanded queries were retrieved using BM25 as this lexical ranker probed to be strong on this task \cite{zhao2023large}. Evaluation results using standard IR metrics are presented in Table \ref{IRtask}. For computational reasons, we evaluated the expansion using a 4-bit quantized version of PatientDx 8B and limit tokens generation size to 200. The rank fusion with BM25 trough RRF was also performed using \citet{BassaniECIR22}. Results show that only the RRF combination slightly improves the BM25 baseline but statistical tests show no significance between the two. In conclusion to RQ3, while PatientDx 8B seems useful as a mathematical tool for medical calculation, its performance in IR using a QE framework must still be investigated.

\begin{table}[]
\begin{center}
\resizebox{\columnwidth}{!}{%
\begin{tabular}{lccccc}
\hline
Model        & MRR@1000 & P@10   & NDCG@10 & Recall@1000 & MAP@100 \\ \hline
(a) BM25 - No QE         & 0.192    & \textbf{0.043}  & 0.154   & 0.756       & 0.128   \\
(b) QE with PatientDx 8B & 0.189    & 0.042  & 0.152   & 0.755       & 0.126   \\
RRF on (a) and (b)          & \textbf{0.193} & \textbf{0.043} & \textbf{0.156}  & \textbf{0.759}      & \textbf{0.129}  \\ \hline
\end{tabular} 
}
\vspace{-0.2cm}
\caption{Retrieval performances of the LLMs in a similar patients task. Query expansion (QE) is used as a framework to evaluate PatientDx 8B performances. \label{IRtask}}

\end{center}

\end{table}

\section{Conclusion and Future Work}
In this paper, we studied the merging of LLMs as a competitive strategy to obtain new sharable models with competitive prediction capabilities and no risks of data privacy violation. Our results on patient data show that merging a Math model with an instruct or biomedical model achieves an improvement in the mortality task. As a major observation, we can highlight an outstanding improvement of 7\% when comparing PatientDx 8B against input LLMs. Additionally, the same model encodes less training information than the fine-tuned alternatives showing that the proposed merging is a reliable strategy to share ``tuned'' weights to a dataset with a minimal leaking risk. Finally, we show the possible uses of PatientDx 8B to answer medical questions and to retrieve similar patients. Despite the advances in this paper, some limitations are still present. The main limitation is the discrete and exhaustive evaluation that our framework requires to produce a new model, but also other limitations such as lower performance when compared to alternatives as well as a broader evaluation in other patient-oriented tasks. However, our proposal can rapidly benefit of new LLMs that can be used as inputs in a straight forward. Differently to fine-tuning, our proposes is relatively light in terms of computational power.   
Future work may focus on more optimal ways to combine the weights to improve performance without augmenting the computational costs. Works such as \citet{akiba2024evolutionary} may be an interesting way to explore more complex merging strategies.
\section*{Limitations}

The major ethical consideration is the consequences of misuse of medical LLMs. Note that this work is intended for use in an academic environment and to support the medical workforce and research\footnote{For any medical concern, please consult a specialist.}. In order to evaluate the generalization capabilities of our model, hyper-parameter selection could be performed on the full training set (without $k$-fold on test as described in §\ref{sec:dataset}) but at significantly higher computational cost. 

\section*{Acknowledgments}
This work has been supported by the In-Utero project funded by HDH (France) and FRQS (Canada).

\bibliography{acl_latex}

\appendix



\end{document}